\documentclass{article}

\usepackage[preprint]{neurips_2025}

\usepackage[utf8]{inputenc} 
\usepackage[T1]{fontenc}    
\usepackage{hyperref}       
\usepackage{url}            
\usepackage{booktabs}       
\usepackage{amsfonts}       
\usepackage{nicefrac}       
\usepackage{microtype}      
\usepackage{inconsolata}
\usepackage{caption}
\usepackage{amsmath}
\usepackage{amssymb}
\usepackage{mathtools}
\usepackage{enumitem}
\usepackage{makecell} 
\usepackage[usestackEOL]{stackengine}
\usepackage{graphicx}
\usepackage{capt-of}
\usepackage{booktabs}
\usepackage{varwidth}
\usepackage[table,dvipsnames,svgnames]{xcolor}

\usepackage{tikz}
\usetikzlibrary{shadows}
\usepackage{float}
\usepackage{caption}
\usepackage{subcaption}
\usepackage{xspace}
\usepackage{svg}

\usepackage{import}
\usepackage{animate}
\usepackage{arydshln}
\usepackage{multirow}
\usepackage[normalem]{ulem}
\usepackage[most]{tcolorbox}
\usepackage{colortbl}
\usepackage{tcolorbox}
\usepackage{pifont}
\usepackage{alltt}
\definecolor{titlegray}{rgb}{0.4, 0.4, 0.4} 
\definecolor{contentgray}{rgb}{0.95, 0.95, 0.95} 
\usepackage{algorithm}
\usepackage{algpseudocode}
\usepackage{todonotes}

\definecolor{carolinablue}{rgb}{0.6, 0.73, 0.89}
\definecolor{mildgreen}{rgb}{0.85, 0.98, 0.80}
\definecolor{beautycolor}{rgb}{0.91, 0.75, 0.96} 
\definecolor{fallacycolor}{rgb}{0.85, 0.95, 1}
\definecolor{gendercolor}{rgb}{1, 0.85, 0.85}
\definecolor{brightyellow}{RGB}{255, 255, 100}
\definecolor{boxcolor}{RGB}{51,51,153}
\definecolor{lightgreen}{rgb}{0.56, 0.93, 0.56}
\definecolor{citeblue}{HTML}{0064E0}
\definecolor{rowhighlight}{RGB}{209, 235, 241}

\hypersetup{
    colorlinks=true,
    citecolor=citeblue,
    linkcolor=red, 
    urlcolor=citeblue 
}

\definecolor{deepblue}{RGB}{0, 0, 139}

\title{Length-Unbiased Sequence Policy Optimization: 

Revealing and Controlling Response Length Variation in RLVR}

\newtcolorbox{questionbanner}{
  colback=blue!10!white,    
  colframe=blue!80!black,   
  width=\textwidth,
  arc=4mm,                  
  boxrule=1pt,              
  fonttitle=\bfseries,
  title=Question,
}
\newtcolorbox{promptbox}[1]{
  colback=contentgray,      
  colframe=titlegray,       
  colbacktitle=titlegray,   
  coltitle=white,           
  title={#1}, 
  arc=4mm,                  
  rounded corners=northwest, 
  rounded corners=northeast, 
  sharp corners=south,      
  boxrule=1pt,              
  fonttitle=\bfseries,      
}

\definecolor{questionbg}{RGB}{240, 248, 255}  
\definecolor{answerbg}{RGB}{245, 255, 250}   
\definecolor{bordercolor}{RGB}{100, 149, 237} 
\definecolor{titlecolor}{RGB}{25, 25, 112}    
\definecolor{rowaccent}{RGB}{235, 235, 235}

\newtcolorbox{vqaexample}[2][]{
    enhanced,
    breakable,
    colback=white,
    colframe=bordercolor,
    boxrule=1.5pt,
    arc=4pt,
    outer arc=4pt,
    left=8pt,
    right=8pt,
    top=8pt,
    bottom=8pt,
    drop shadow={shadow xshift=0.5mm, shadow yshift=-0.5mm, opacity=0.3},
    overlay={
        \node[
            anchor=north east,
            xshift=-3pt,
            yshift=-3pt,
            fill=bordercolor!80,
            text=white,
            font=\bfseries,
            rounded corners=2pt,
            inner sep=4pt,
            minimum height=1.2em,
            align=center
        ] at (frame.north east) {#2};
    },
    #1
}

\author{
    Fanfan Liu \thanks{Equal contribution} ,\hspace{0.3em}
    Youyang Yin \footnotemark[1] ,\hspace{0.3em}
    Peng Shi ,\hspace{0.3em}
    Siqi Yang ,\hspace{0.3em}
    Zhixiong Zeng ,\hspace{0.3em}
    Haibo Qiu\hspace{0.3em}\\
    \\
    \textbf{Meituan} \\ 
}

\begin{document}

\maketitle
\vspace{-5mm}

\begin{abstract}
Recent applications of Reinforcement Learning with Verifiable Rewards (RLVR) to Large Language Models (LLMs) and Vision-Language Models (VLMs) have demonstrated significant success in enhancing reasoning capabilities for complex tasks. During RLVR training, an increase in response length is often regarded as a key factor contributing to the growth of reasoning ability. However, the patterns of change in response length vary significantly across different RLVR algorithms during the training process. To provide a fundamental explanation for these variations, this paper conducts an in-depth analysis of the components of mainstream RLVR algorithms. We present a theoretical analysis of the factors influencing response length and validate our theory through extensive experimentation. Building upon these theoretical findings, we propose the Length-Unbiased Sequence Policy Optimization (LUSPO) algorithm. Specifically, we rectify the length bias inherent in Group Sequence Policy Optimization (GSPO), rendering its loss function unbiased with respect to response length and thereby resolving the issue of response length collapse. We conduct extensive experiments across mathematical reasoning benchmarks and multimodal reasoning scenarios, where LUSPO consistently achieves superior performance. Empirical results demonstrate that LUSPO represents a novel, state-of-the-art optimization strategy compared to existing methods such as GRPO and GSPO.
\end{abstract}



\section{Introduction}

Reinforcement Learning with Verifiable Rewards (RLVR) has become a cornerstone technique for advancing the capabilities of language models \citep{Guo_2025}. Leveraging large-scale RLVR, these models are empowered to address complex tasks, including sophisticated mathematics and high-level programming, through extended and nuanced reasoning. Through iterative optimization and exposure to diverse reward signals, RLVR-trained language models develop the ability to generate longer, more coherent, and contextually relevant responses. This facilitates nuanced reasoning, allowing the models to break down complex problems into manageable steps and provide detailed, logical explanations. As a result, RLVR not only improves the accuracy and reliability of language model outputs, but also significantly expands their applicability across domains requiring sophisticated reasoning and decision-making. 

Among contemporary RLVR strategies, GRPO \citep{shao2024deepseekmathpushinglimitsmathematical} has emerged as a prominent method, demonstrating strong performance in DeepSeek-R1 \citep{Guo_2025}. Nonetheless, GRPO’s improper utilization of importance sampling introduced instability in the training process for Mixture-of-Experts (MoE) architectures. GSPO \citep{zheng2025groupsequencepolicyoptimization} subsequently resolved this limitation by employing sequence-level importance weighting \citep{zheng2023clickcontrollabletextgeneration}, thereby enhancing stability in MoE training.

Although GRPO has achieved notable success, averaging the contribution of all tokens within each trajectory in the GRPO objective leads to a length bias. This bias causes the model to give larger gradient updates to shorter responses, encouraging brevity in correct answers. On the other hand, for negative advantages (i.e., incorrect responses), longer outputs are penalized less, making the policy favor longer responses when the answer is incorrect.

GSPO exhibits the same issue as well, Moreover, the GSPO's objective function further exacerbates this bias. Sequence-level clipping leads to a substantially greater number of tokens being truncated compared to token-level clipping. Additionally, the clip-higher mechanism \citep{yu2025dapoopensourcellmreinforcement} disproportionately removes negative sample tokens, resulting in a pronounced imbalance between positive and negative sample tokens. This disparity amplifies the response-level length bias inherent in the GSPO's objective function, where positive samples incentivize the model to generate shorter outputs. Over time, this bias causes the model to produce increasingly brief responses, undermining overall training efficacy.

\begin{figure}[H]
    \centering
    \includegraphics[width=0.4\textwidth]{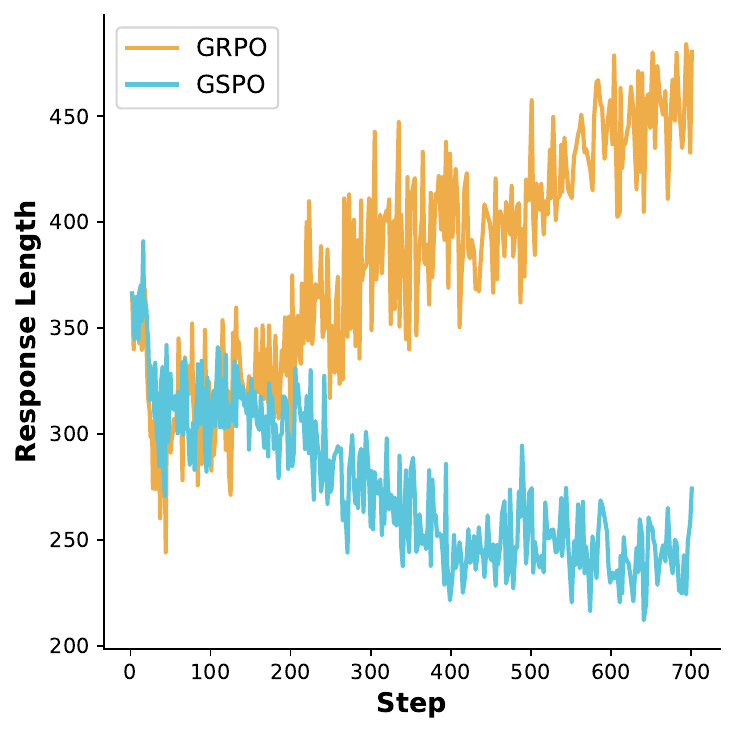}
    \caption{Response length during RLVR training for Qwen2.5-VL-7B-Instruct. Under strictly controlled experimental settings (with all conditions except for the loss function kept constant), we compared the response length curves of GRPO and GSPO. It can be observed that GRPO induces the model to generate longer responses, while GSPO leads the model to gradually shorten its response length during training.}
    \label{fig:response_length_algorithm}

    
\end{figure}


    

As illustrated in the figure \ref{fig:response_length_algorithm}, training Qwen2.5-VL-7B-Instruct \citep{bai2025qwen25vltechnicalreport} under identical experimental settings, GSPO exhibited response length collapse during training, while GRPO did not show this phenomenon.

In this work, we revisit the GRPO's and GSPO's objective function and present a thorough analysis. To eliminate its inherent length bias, we introduce the Length-Unbiased Sequence Policy Optimization (LUSPO) algorithm, which implements a straightforward yet impactful modification: scaling each sequence’s loss by its own length. This adjustment alleviates the degraded performance observed with GSPO in dense models, while maintaining robust training dynamics in MoE models, and significantly accelerates the growth of response length during training and improves performance on reasoning tasks, both in text-only and multimodal settings.

To substantiate the utility of our approach, we perform comprehensive empirical studies across diverse architectures, encompassing dense, MoE, text-only, and vision-language (VL) models. Experimental results show that LUSPO can effectively eliminate the length bias of the GSPO's objective function, ensuring stable training. Moreover, evaluations on a suite of text-only and multimodal benchmarks—including AIME24 \citep{aime24}, AIME25 \citep{aime25}, MathVista \citep{lu2024mathvista}, MathVision \citep{wang2024measuring}, and others—demonstrate notable improvements over GRPO and GSPO. For example, training Qwen2.5-7B-Base \citep{qwen2025qwen25technicalreport} and Qwen3-30B-A3B-Instruct \citep{yang2025qwen3technicalreport} with LUSPO yields up to 2.9\% and 6.9\% higher accuracy on AIME24 compared to GSPO, meanwhile, training Qwen2.5-VL-7B-Instruct \citep{bai2025qwen25vltechnicalreport} with LUSPO results in up to 1.6\% higher accuracy than GRPO and 0.5\% higher accuracy than GSPO on the MathVista-Mini.

In summary, our key contributions are:

\begin{itemize}
    \item We conduct a detailed analysis of the objective functions of GRPO and GSPO, and clarify the underlying reasons for their inherent length bias.
    \item We propose the LUSPO algorithm, which neutralizes GSPO’s length bias through a principled modification.
    \item We conduct extensive experiments across multiple models to validate the effectiveness and generalizability of our method.
\end{itemize}
\section{Related Work}

\subsection{Reinforcement Learning with Verifiable Rewards}
With the tremendous success of DeepSeek-R1 \citep{Guo_2025}, RLVR has become widely adopted for post-training large language models. At the same time, a number of RLVR algorithms have been proposed by researchers. The most representative among them is Group Relative Policy Optimization (GRPO) \citep{shao2024deepseekmathpushinglimitsmathematical}, which eliminates the necessity for a value model by calculating the relative advantage of each response within a group corresponding to the same query. To enhance the effectiveness of reinforcement learning, Decoupled Clip and Dynamic Sampling Policy Optimization (DAPO) \citep{yu2025dapoopensourcellmreinforcement} incorporates four main techniques: Clip-Higher, Dynamic Sampling, Token-Level Policy Gradient Loss, and Overlong Reward Shaping. Moreover, Dr.GRPO \citep{liu2025understanding} proposes an unbiased optimization approach that enhances token efficiency without compromising reasoning performance.

\subsection{RLVR on MoE models}
Currently, an increasing number of leading large models are actively exploring and adopting the MoE architecture. Significant progress is being made in areas such as efficient training, expert routing algorithms, and sparse activation techniques. MoE models are poised to become a foundational architecture for next-generation general-purpose large models, driving major advancements in AI capabilities for complex reasoning and cross-domain integration.The Selection of an appropriate RLVR algorithm for post-training MoE models has become critically important. 

However, GRPO and its various extensions rely on token-level importance ratios, which tend to exhibit high variance—particularly in Mixture-of-Experts (MoE) models, where routing diversity and longer responses further amplify token-level fluctuations. This increased variance makes unstable updates more likely during training. 

To tackle this problem, Group Sequence Policy Optimization (GSPO) \citep{zheng2025groupsequencepolicyoptimization} introduces an importance ratio defined at the sequence level, rather than at the token level, and applies sequence-level clipping, reward assignment, and optimization. GSPO exhibits substantial improvements over GRPO in terms of training stability, efficiency, and overall performance. Importantly, GSPO inherently addresses the stability issues associated with RL training of large Mixture-of-Experts (MoE) models, removing the necessity for intricate stabilization techniques. In addition, Soft Adaptive Policy Optimization (SAPO) \citep{gao2025softadaptivepolicyoptimization} is a token-adaptive and smooth reinforcement learning algorithm developed to mitigate the instability and inefficiency found in hard-clipped policy optimization for large language models. Instead of relying on discontinuous clipping, SAPO utilizes a temperature-controlled soft gating mechanism, and applies asymmetric temperatures to more effectively manage negative-token gradients. This approach delivers a more stable and informative optimization signal throughout training.
\section{Preliminaries}

\paragraph{Notation}

In this work, we model an autoregressive language model parameterized by $\theta$ as a policy $\pi_\theta$. Let $x$ denote a query and $\mathcal{D}$ be the query set. Given a response $y$ to an input query $x$, its likelihood under the policy $\pi_\theta$ is expressed as $\pi_\theta (y | x)=\prod_{t=1}^{|y|} \pi_\theta (y_t | x, y_{<t} )$ where $|y|$ represents the number of tokens in $y$. Each query-response pair $(x, y)$ can be evaluated by a verifier $r$, which assigns a reward value $r(x, y)$ to the pair.

\subsection{GRPO: Group Relative Policy Optimization}
For each query $x \in \mathcal{D}$, GRPO \citep{shao2024deepseekmathpushinglimitsmathematical} samples a group of responses from the behavior policy, computes their rewards and optimizes the following objective:


\begin{equation}
\resizebox{\textwidth}{!}{
$\mathcal{J}_\text{GRPO}(\theta) = \mathbb{E}_{ x \sim \mathcal{D},\, \{y_i\}_{i=1}^G 
    \sim \pi_{\theta_\text{old}}( \cdot | x) }
\left[
    \frac{1}{G} \sum_{i=1}^{G} \frac{1}{|y_i|} \sum_{t=1}^{|y_i|} 
    \min \left( w_{i,t}(\theta) \widehat{A}_{i,t},
    \mathrm{clip} \left( w_{i,t}(\theta), 1 - {\varepsilon}, 1 + {\varepsilon}\right) \widehat{A}_{i,t} \right)
\right]$
}
\label{equ:grpo}
\end{equation}

where $G$ denotes the number of responses generated for each query $x$ (i.e., the size of the group), and the importance ratio $w_{i,t}(\theta)$ and advantage $\widehat{A}_{i,t}$ of token $y_{i,t}$ are defined as follows:
\begin{align}
    w_{i,t}(\theta)=\frac{ \pi_{\theta} (y_{i,t} | x, y_{i,<t}) }{ \pi_{\theta_\text{old}} (y_{i,t} | x,y_{i,<t})},
\end{align}
    
\begin{align}
    \widehat{A}_{i,t} = \widehat{A}_{i} = \frac{r(x, y_i) - \mathrm{mean} \left( \{ r(x, y_i) \}_{i=1}^G \right) }{ \mathrm{std} \left( \{ r(x, y_i) \}_{i=1}^G \right) },
\end{align}
respectively, with all tokens in $y_i$ sharing a common advantage given by $\widehat{A}_{i}$.

\subsection{GSPO: Group Sequence Policy Optimization}
The use of the token-level importance weigh $\frac{ \pi_{\theta} (y_{i,t} | x, y_{i,<t}) }{ \pi_{\theta_\text{old}} (y_{i,t} | x,y_{i,<t})}$ in GRPO presents certain issues. In contrast, the sequence-level importance weight $\frac{ \pi_{\theta} (y | x) }{ \pi_{\theta_\text{old}} (y | x)}$ is theoretically well-founded in the context of language generation: it quantifies the extent to which a response $y$ sampled from $\pi_{\theta_\text{old}} (\cdot | x)$ differs from the current policy $\pi_{\theta} (\cdot | x)$. This measure is naturally compatible with sequence-level rewards and provides an intuitive basis for the clipping mechanism.

Building on this intuitive insight, GSPO \citep{zheng2025groupsequencepolicyoptimization} adopts the following sequence-level optimization objective:


\begin{equation}
\mathcal{J}_\text{GSPO} (\theta) =
\mathbb{E}_{ x \sim \mathcal{D},\, \{y_i\}_{i=1}^G 
    \sim \pi_{\theta_\text{old}}( \cdot | x) }
\left[
    \frac{1}{G} \sum_{i=1}^{G}
    \min \left( s_{i}(\theta) \widehat{A}_{i},
    \mathrm{clip} \left( s_{i}(\theta), 1 - {\varepsilon}, 1 + {\varepsilon} \right) \widehat{A}_{i} \right)
\right]
\label{equ:gspo}
\end{equation}

where the group-based advantage is estimated as
\begin{align}
\widehat{A}_{i} = \frac{r(x, y_i) - \mathrm{mean} \left( { r(x, y_i) }_{i=1}^G \right) }{ \mathrm{std} \left( { r(x, y_i) }_{i=1}^G \right) },
\end{align}
and the importance ratio $s_{i}(\theta)$ is defined according to the sequence likelihood \citep{zheng2023clickcontrollabletextgeneration}:

\begin{equation}
s_{i}(\theta) = \left( \frac{ \pi_{\theta} (y_i | x) }{ \pi_{\theta_\text{old}} (y_i | x)} \right)^{\frac{1}{|y_i|}}
= \exp \left( \frac{1}{|y_i|} \sum_{t=1}^{|y_i|} \log \frac{ \pi_{\theta} (y_{i,t} | x, y_{i,<t}) }{ \pi_{\theta_\text{old}} (y_{i,t} | x,y_{i,<t})} \right).
\end{equation}
\section{Algorithm}


Response length is a key metric in RLVR for large language models. By extending the response length, the model is able to explore a wider range of reasoning patterns and develop more advanced problem-solving strategies throughout the training process.

\subsection{Analysis of Response Length Variation}

There are many factors that influence changes in response length during training, which mainly fall into two aspects.

\begin{figure}[H]
    \centering
    \includegraphics[width=0.4\textwidth]{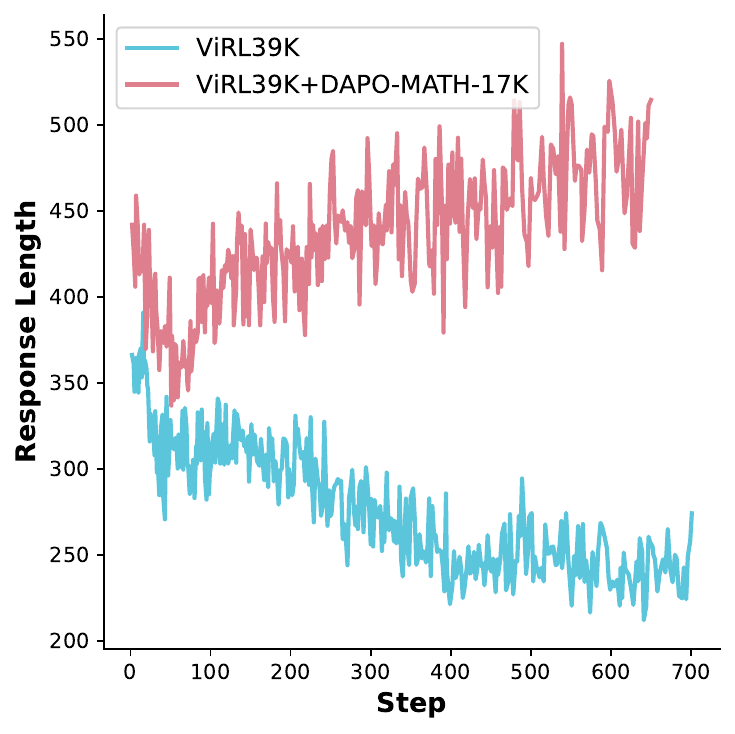}
    \caption{Response length during the training of Qwen2.5-VL-7B-Instruct with GSPO on different datasets exhibits different trends.}
    \label{fig:response_length_data}

    
\end{figure}

One aspect is the reward-driven changes in response length. For example, explicitly adding incentives or penalties for response length in the reward function can directly encourage longer or shorter responses. Alternatively, implicit effects may occur: if longer answers are more likely to receive correct rewards, the model tends to generate longer responses; if shorter answers are favored, the response length decreases accordingly, this may be related to the choice of training data. As illustrated in the figure \ref{fig:response_length_data}, training Qwen2.5-VL-7B-Instruct on ViRL39k \citep{vl-rethinker} and DAPO-MATH-17K \citep{yu2025dapoopensourcellmreinforcement} led to a growth in response length, while only training on ViRL39k caused response length collapse. This is because DAPO-MATH-17K requires relatively more complex reasoning processes.

Another aspect, and also the main focus of this paper, is the preference for response length embedded in the loss function, as exemplified by approaches like GRPO \cite{shao2024deepseekmathpushinglimitsmathematical} and GSPO \citep{zheng2025groupsequencepolicyoptimization}. An in-depth analysis will be presented in the subsequent subsections.

\subsection{Response Length Bias in GSPO}

In the GRPO's loss function, the inner layer first computes the average over tokens within each trajectory, and the outer layer then averages across all trajectories. This calculation leads to the following issue:

For long responses, since the number of tokens in a trajectory is large, dividing by the trajectory length means each token in a long response contributes less to the loss, while for short responses, each token contributes more.

If the sample is correct, tokens in short responses have a higher weight, so the model tends to generate shorter responses. If the sample is incorrect, the model avoids very short responses and favors longer responses. Thus, step accuracy influences response length: high accuracy leads to shorter responses, while low accuracy leads to longer ones.

In a single step of the GRPO algorithm, the contributions to the gradient from positive and negative samples are generally balanced, the impact of this length bias on training is not very significant. However, this length bias has a more pronounced impact on training process of GSPO.

Although GSPO's loss function replaces token-level importance ratios with an importance ratio based on sequence likelihood, it still does not address the aforementioned issue of length bias.

At the same time, GSPO adopts sequence-level clipping, which results in a significant increase in the proportion of tokens being clipped compared to GRPO. Additionally, in practical applications, GSPO applies the Clip-Higher operation, causing the number of clipped negative samples to exceed that of positive samples. This indirectly leads to the gradients in a single step being dominated by positive samples, making the model's responses tend to become increasingly shorter.

\subsection{Length-Unbiased Sequence Policy Optimization}

To address this issue, we proposed Length-Unbiased Sequence Policy Optimization (LUSPO). LUSPO introduces a simple yet effective modification to the GSPO's objective by scaling each sequence’s loss according to its own length. This adjustment directly mitigates the response-level length bias inherent in GSPO, ensuring that longer sequences are not unfairly penalized during training. As a result, LUSPO not only alleviates the collapse in response length observed with GSPO, but also promotes more balanced and stable learning dynamics across both dense and Mixture-of-Experts (MoE) architectures. 

LUSPO’s optimization objective is as follows:


\begin{equation}
\mathcal{J}_\text{LUSPO} (\theta) =
\mathbb{E}_{ x \sim \mathcal{D},\, \{y_i\}_{i=1}^G \sim \pi_{\theta_\text{old}}( \cdot | x) }
\left[
    \frac{1}{G} \sum_{i=1}^{G}
    \min \left( s_{i}(\theta) \widehat{A}_{i}, 
    \mathrm{clip}\left( s_{i}(\theta), 1 - {\varepsilon}, 1 + {\varepsilon} \right) \widehat{A}_{i} \right) \cdot |y_i|
\right]
\label{equ:luspo}
\end{equation}

where the group-based advantage estimation and importance ratio take the same form as in GSPO.


\subsection{Gradient Analysis}

The gradient of the LUSPO objective can be derived as follows (with clipping omitted for clarity):


{\fontsize{8.5pt}{10.5pt}\selectfont
\begin{align}
\nabla_{\theta} \mathcal{J}_\text{LUSPO} (\theta)
=&\
\nabla_{\theta} \mathbb{E}_{ x \sim \mathcal{D},\, \{y_i\}_{i=1}^G 
    \sim \pi_{\theta_\text{old}}( \cdot | x) } \left[
    \frac{1}{G} \sum_{i=1}^{G} s_{i}(\theta) \widehat{A}_{i} \cdot |y_i|
\right] \nonumber \\
=&\
\mathbb{E}_{ x \sim \mathcal{D},\, \{y_i\}_{i=1}^G 
    \sim \pi_{\theta_\text{old}}( \cdot | x) } \left[
    \frac{1}{G} \sum_{i=1}^{G}
    s_{i}(\theta) \widehat{A}_{i} \cdot |y_i| 
    \cdot \nabla_{\theta} \log s_{i}(\theta)
\right] \nonumber \\
=&\
\mathbb{E}_{ x \sim \mathcal{D},\, \{y_i\}_{i=1}^G 
    \sim \pi_{\theta_\text{old}}( \cdot | x) } \Bigg[
    \frac{1}{G} \sum_{i=1}^{G}
    \left( \frac{ \pi_{\theta} (y_i | x) }
           { \pi_{\theta_\text{old}} (y_i | x)} \right)^{\frac{1}{|y_i|}}
    \widehat{A}_{i} \cdot |y_i|
    \cdot \frac{1}{|y_i|} \sum_{t=1}^{|y_i|} 
    \nabla_{\theta} \log \pi_{\theta} (y_{i,t} | x, y_{i,<t})
\Bigg] \nonumber \\
=&\
\mathbb{E}_{ x \sim \mathcal{D},\, \{y_i\}_{i=1}^G 
    \sim \pi_{\theta_\text{old}}( \cdot | x) } \Bigg[
    \frac{1}{G} \sum_{i=1}^{G}
    \left( \frac{ \pi_{\theta} (y_i | x) }
           { \pi_{\theta_\text{old}} (y_i | x)} \right)^{\frac{1}{|y_i|}}
    \widehat{A}_{i}
    \sum_{t=1}^{|y_i|} 
    \nabla_{\theta} \log \pi_{\theta} (y_{i,t} | x, y_{i,<t})
\Bigg].
\label{equ:luspo_grad}
\end{align}
}

Similarly, we can derive the gradient of the GSPO objective as follows:


\begin{equation}
\resizebox{\textwidth}{!}{
$\nabla_{\theta} \mathcal{J}_\text{GSPO} (\theta) =
\mathbb{E}_{ x \sim \mathcal{D},\, \{y_i\}_{i=1}^G \sim \pi_{\theta_\text{old}}( \cdot | x) }
\left[
    \frac{1}{G} \sum_{i=1}^{G}
    \left( \frac{ \pi_{\theta} (y_i | x) }{ \pi_{\theta_\text{old}} (y_i | x)} \right)^{\frac{1}{|y_i|}}
    \widehat{A}_{i}
    \frac{1}{|y_i|} \sum_{t=1}^{|y_i|} 
    \nabla_{\theta} \log \pi_{\theta} (y_{i,t} | x, y_{i,<t})
\right].$
}
\label{equ:gspo_grad}
\end{equation}

By comparing the gradients of the LUSPO and GSPO objectives, it can be observed that LUSPO eliminates the length-dependent bias present in GSPO for each trajectory.
\section{Experiment}
\label{section:experiment}

\subsection{Training Setup}
In this work, to assess the generalizability of our method, we performed comprehensive experiments across both dense and Mixture-of-Experts (MoE) architectures, as well as on text-only and vision-language (VL) models.

\subsubsection*{Implement Details}

For model backbones, we utilize Qwen2.5-7B-Base \citep{qwen2025qwen25technicalreport} to represent dense models and Qwen3-30B-A3B-Instruct \citep{yang2025qwen3technicalreport} to represent MoE models, both for text-only tasks. For multimodal evaluations, Qwen2.5-VL-7B-Instruct \citep{bai2025qwen25vltechnicalreport} serves as the model for training. 

We trained Qwen2.5-7B-Base and Qwen2.5-VL-7B-Instruct on 8 Nvidia H800 GPUs and trained Qwen3-30B-A3B-Instruct on 4 x 8 Nvidia H800 GPUs. We trained these models mainly on verl \citep{sheng2024hybridflow} framework.


Regarding hyperparameters, we adopt the AdamW \citep{loshchilov2019decoupledweightdecayregularization} optimizer with a fixed learning rate of $1 \times 10^{-6}$, coupled with a linear warm-up \citep{vaswani2017attention} spanning 20 rollout steps. Each prompt batch comprises 128 items, with 8 responses sampled per prompt during rollout. Training utilizes a mini-batch size of 16, and the maximum generation length is set to 32,768 and 4,096 tokens for text-only and VL models, respectively. For the Clip-Higher mechanism \citep{yu2025dapoopensourcellmreinforcement}, we configure $\epsilon_{low}$ to $2 \times 10^{-3}$ and $\epsilon_{high}$ to $2.5 \times 10^{-3}$, striking an effective balance between exploration and exploitation. The top-p is set to 0.7 and the temperature is set to 1.0 for the actor rollout.

\subsubsection*{Dataset}

\vspace{-1em}

\begin{table}[H]
\centering
\caption{Description of datasets.}
\label{tab:dataset}
\renewcommand{\arraystretch}{1.1} 
\setlength{\tabcolsep}{0pt}       

\begin{tabular}{c c}
\toprule
\textbf{Dataset} & \textbf{Description}\\
\midrule

\multirow{2}{*}{DAPO-17K-MATH} & math questions paired with  \\ 
                            & corresponding single integer answers \\
\midrule

\multirow{4}{*}{ViRL39K} & Math/Phys/Chem/Bio \\
                      & charts/diagrams/tables-based reasoning \\
                      & broader STEM \\
                      & social science topics \\

\bottomrule
\end{tabular}
\end{table}

\vspace{-1em}

We trained on the DAPO-MATH-17K \citep{yu2025dapoopensourcellmreinforcement} dataset, while the VL model was trained on the ViRL39K \citep{vl-rethinker} dataset. As shown in table \ref{tab:dataset}, both datasets focus on scientific-related problems. Therefore, our primary evaluation focuses on mathematical and logical tasks, which serve as robust testbeds for our algorithm and can be seamlessly extended to other domains.

\subsubsection*{Reward}

The reward we designed during training consists of three components, taking into account accuracy, format, and response length.

\begin{align}
\mathcal{R} = \mathcal{R}_{\mathit{accuracy}} + \mathcal{R}_{\mathit{format}} + \mathcal{R}_{\mathit{overlong}}
\end{align}

\begin{figure}[H]

    \centering
    \includegraphics[width=0.8\linewidth,clip,trim=120 525 120 80]{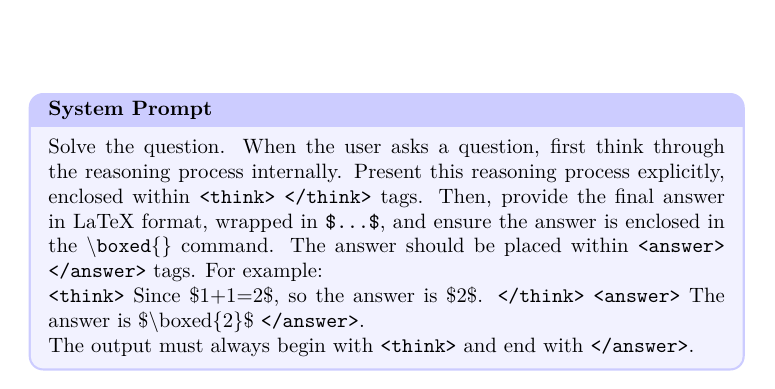}
    \caption{System prompt used during VL model training}
    \label{fig:prompt}
\end{figure}

The accuracy reward $\mathcal{R}_{\mathit{accuracy}} \in \{0, 1\}$, depending on whether the answer is correct. The format reward $\mathcal{R}_{\mathit{format}} \in \{0, 0.5\}$, depending on whether the required format specified in the prompt is followed (figure \ref{fig:prompt}).

\vspace{-2em}

The overlong reward is defined as follows:

\begin{align}
\mathcal{R}_{\mathit{overlong}}(y) = 
\min\left\{ 0, \; \frac{ \left( L_{\mathit{max}} - L_{\mathit{buffer}} \right) - |y| }{ L_{\mathit{buffer}} } \times 0.3 \right\}
\end{align}

where $L_{\mathit{max}}$ is the maximum generation length, $L_{\mathit{buffer}}$ is set to 512 for Qwen2.5-VL-7B-Instruct and 4096 for Qwen2.5-7B-Base and Qwen3-30B-A3B-Instruct and $|y|$ is the length of response.



    

\subsection{Main Results}


We adopt GRPO and GSPO as our primary baseline algorithms for comparison. For text-only models, we evaluated their performance on benchmarks, including AMC23, AIME24 \citep{aime24}, AIME25 \citep{aime25} and MATH500 \citep{lightman2023letsverifystepstep}. For Qwen3-30B-A3B-Instruct, due to it's powerful capabilities, we conduct evaluations only on AIME.

\begin{table}[H]

\centering
\caption{Comparison of GSPO and LUSPO-trained Qwen2.5-7B-Base and Qwen3-30B-A3B-Instruct models on text-only benchmarks.}
\label{tab:text-only_benchmark}
\renewcommand{\arraystretch}{1.1} 
\setlength{\tabcolsep}{17pt}       
\resizebox{\textwidth}{!}{
\begin{tabular}{l c c c c c}
\toprule
\textbf{Base model + Algorithm} & AMC23 & AIME24 & AIME25 & MATH500 & \textbf{Avg.} \\
\midrule

Qwen2.5-7B-Base & & & & &\\
\hspace{1em}w/o RLVR      & 35.8 & 1.6 & 3.6 & 60.8 & 25.5 \\
\hspace{1em}GSPO      & 55.3 & 11.8 & 11.2 & 71.0 & 37.3 \\
\rowcolor{rowhighlight}
\hspace{1em}LUSPO        & \textbf{58.3} & \textbf{14.7}  & \textbf{13.9} & \textbf{78.4} & \textbf{41.3} \\
\rowcolor{rowaccent}
\hspace{1em}$\Delta$ \textsubscript{GSPO} & +3.0 & +2.9 & +2.7 & +7.4 & +4.0 \\

\midrule

Qwen3-30B-A3B-Instruct & & & & & \\
\hspace{1em}w/o RLVR      & 97.5 & 60.0 & 57.2 & 96.2 & --- \\
\hspace{1em}GSPO      & --- & 76.7 & 59.2 & --- & 68.0 \\
\rowcolor{rowhighlight}
\hspace{1em}LUSPO        & --- & \textbf{83.6} & \textbf{76.3} & --- & \textbf{80.0} \\
\rowcolor{rowaccent}
\hspace{1em}$\Delta$ \textsubscript{GSPO} & --- & +6.9 & +17.1 & --- & +12.0 \\

\bottomrule
\end{tabular}
}
\end{table}

Notably, as shown in table \ref{tab:text-only_benchmark} both the dense model (Qwen2.5-7B-Base) and the Mixture-of-Experts (MoE) model (Qwen3-30B-A3B-Instruct) demonstrated significant improvements.

\begin{table}[H]
\centering
\caption{Comparison of GRPO, GSPO and LUSPO-trained Qwen2.5-VL-7B-Instruct models on multimodal benchmarks.}
\label{tab:vl_benchmark}
\renewcommand{\arraystretch}{1.1} 
\setlength{\tabcolsep}{8pt}       
\resizebox{\textwidth}{!}{
\begin{tabular}{l c c c c c c c}
\toprule
\textbf{Base model + Algorithm} & MathVista-mini & MathVision & MathVerse(Vision Only) & DynaMath & WeMath & LogicVista & \textbf{Avg.} \\
\midrule

Qwen2.5-VL-7B-Instruct &  &  &  &  &  &  & \\
\hspace{1em}w/o RLVR      & 67.4 & 26.2 & 41.1 & 20.2 & 34.5 & 45.6 & 39.2 \\
\hspace{1em}GRPO        & 72.8 & \textbf{28.7} & 46.8 & \textbf{26.2} & 43.3 & 46.5 & 44.0 \\
\hspace{1em}GSPO      & 73.9 & 27.7 & 45.6 & 25.9 & 40.7 & 47.7 & 43.6 \\
\rowcolor{rowhighlight}
\hspace{1em}LUSPO        & \textbf{74.4} & 28.0 & \textbf{47.1} & 24.6 & \textbf{45.6} & \textbf{53.7} & \textbf{45.6} \\
\rowcolor{rowaccent}
\hspace{1em}$\Delta$ \textsubscript{GRPO} & +1.6 & -0.7 & +0.3 & -1.6 & +2.5 & +7.2 & +1.6 \\
\rowcolor{rowaccent}
\hspace{1em}$\Delta$ \textsubscript{GSPO} & +0.5 & +0.3 & +1.5 & +0.7 & +5.1 & +6.0 & +2.0 \\
\bottomrule
\end{tabular}
}
\end{table}

In addition to evaluating text-only models, we further assessed the effectiveness of the proposed LUSPO algorithm in the multimodal domain. Table~\ref{tab:vl_benchmark} presents a comparative analysis of LUSPO, GRPO and GSPO on various multimodal benchmarks, including MathVista-mini \citep{lu2024mathvista}, MathVision \citep{wang2024measuring}, MathVerse(Vision Only) \citep{zhang2024mathverse}, DynaMath \citep{zou2024dynamathdynamicvisualbenchmark}, WeMath \citep{qiao2025we} and LogicVista \citep{xiao2024logicvistamultimodalllmlogical}, using the Qwen2.5-VL-7B-Instruct model. The results demonstrate that LUSPO consistently outperforms GRPO and GSPO, indicating its strong generalization capability across both textual and visual-language tasks. Especially on the Wemath and LogicVista benchmarks, LUSPO achieves improvements of 5.1\% and 6.0\% over GSPO, respectively.

It is worth noting that in evaluations on several multimodal benchmarks, GSPO even achieves lower average scores than GRPO. This is because, as previously discussed, GSPO further amplifies the length bias arises from the operation of averaging the gradient contributions of each token within a trajectory.

\subsection{Training Dynamics}

\subsubsection*{Response Length} 

\begin{figure*}[!htbp]
    \centering

    \begin{subfigure}[b]{0.31\textwidth}
        \centering
        \includegraphics[width=\textwidth]{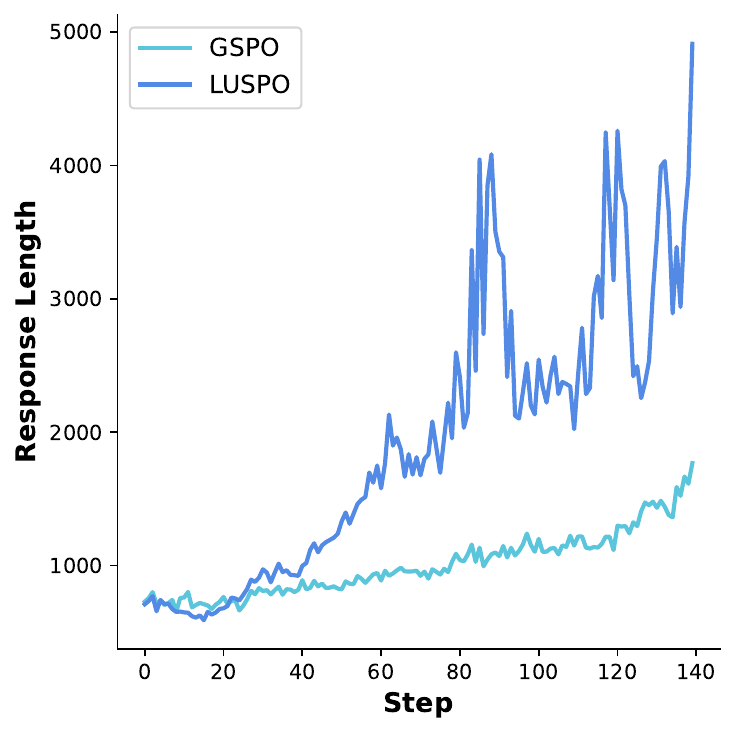}
        \caption{Qwen2.5-7B-Base}
        \label{fig:response_length_qwen2.5-7b-base}
    \end{subfigure}
    \hspace{0.02\textwidth}
    \begin{subfigure}[b]{0.31\textwidth}
        \centering
        \includegraphics[width=\textwidth]{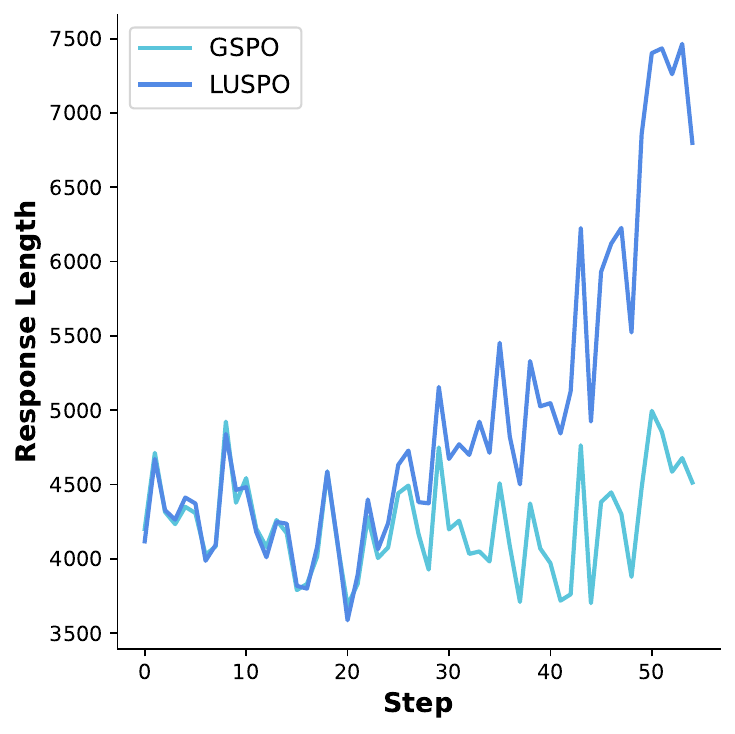}
        \caption{Qwen3-30B-A3B-Instruct}
        \label{fig:response_length_qwen3-30b-a3b-instruct}
    \end{subfigure}
    \hspace{0.02\textwidth}
    \begin{subfigure}[b]{0.31\textwidth}
        \centering
        \includegraphics[width=\textwidth]{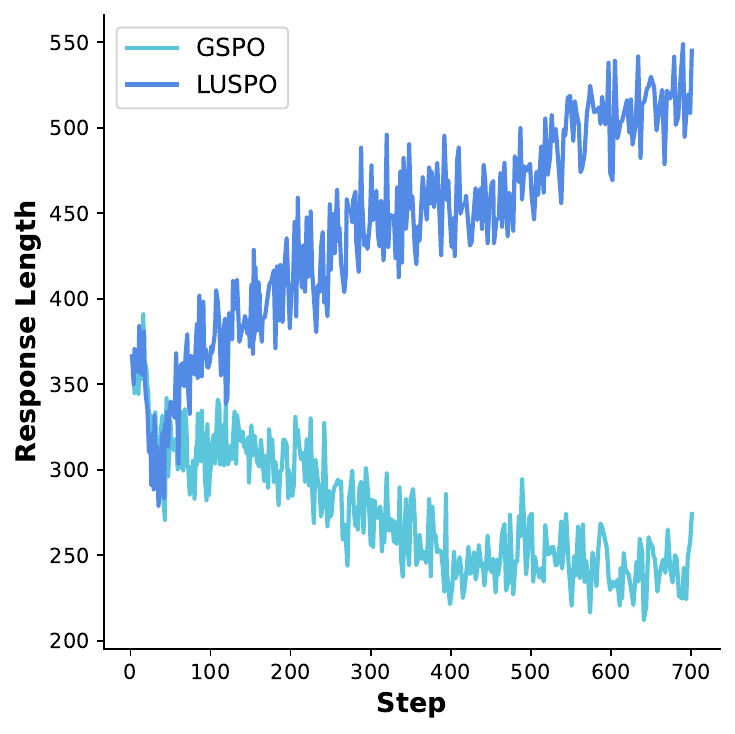}
        \caption{Qwen2.5-VL-7B-Instruct}
        \label{fig:response_length_qwen2.5-vl-7b-instruct}
    \end{subfigure}

    
    \caption{Training curves of GSPO and LUSPO across reponse length.}
    \label{fig:response_length}
\end{figure*}

As illustrated in figure~\ref{fig:response_length}, given the same number of training steps, LUSPO exhibits a more rapid increase in response length compared to GSPO. This accelerated growth in response length is indicative of enhanced model capability, further underscoring the effectiveness of the LUSPO algorithm. 

In particular, as shown in figure~\ref{fig:response_length_qwen2.5-vl-7b-instruct}, within the VL model, GSPO results in a pronounced collapse of response length, which severely restricts the model’s capacity for exploration and complex reasoning. By contrast, LUSPO effectively mitigates this issue by maintaining a stable and sufficiently long response length throughout training. This stability allows the model to better leverage multimodal information and sustain its reasoning ability. This phenomenon of response length collapse also corroborates our earlier analysis that GSPO exacerbates length bias.

\begin{table}[H]
\centering
\caption{Average length of generated responses on validation set.}
\label{tab:response_length}
\renewcommand{\arraystretch}{1.1} 
\setlength{\tabcolsep}{5pt}       

\begin{tabular}{c c c}
\toprule
\multirow{2}{*}{\textbf{Base Model}} & \multicolumn{2}{c}{\textbf{Response Length}} \\
\cmidrule(lr){2-3}
                    & \textbf{GSPO} & \textbf{LUSPO} \\
\midrule

Qwen2.5-7B-Base     & 2611 & 3940\\ 
Qwen3-30B-A3B-Instruct & 6757 & 11014\\


\bottomrule
\end{tabular}
\end{table}

Table \ref{tab:response_length} presents the average response length for GSPO and LUSPO on validation set. For both models the average response length of LUSPO is nearly 1.5 times that of GSPO.

\subsubsection*{Accuracy Reward}

Accuracy reward obtained during training has long been recognized as a fundamental metric for monitoring the progress and effectiveness of reinforcement learning algorithms. 

\begin{figure}[H]
    \centering

    \begin{subfigure}[b]{0.3\textwidth}
        \centering
        \includegraphics[width=\linewidth]{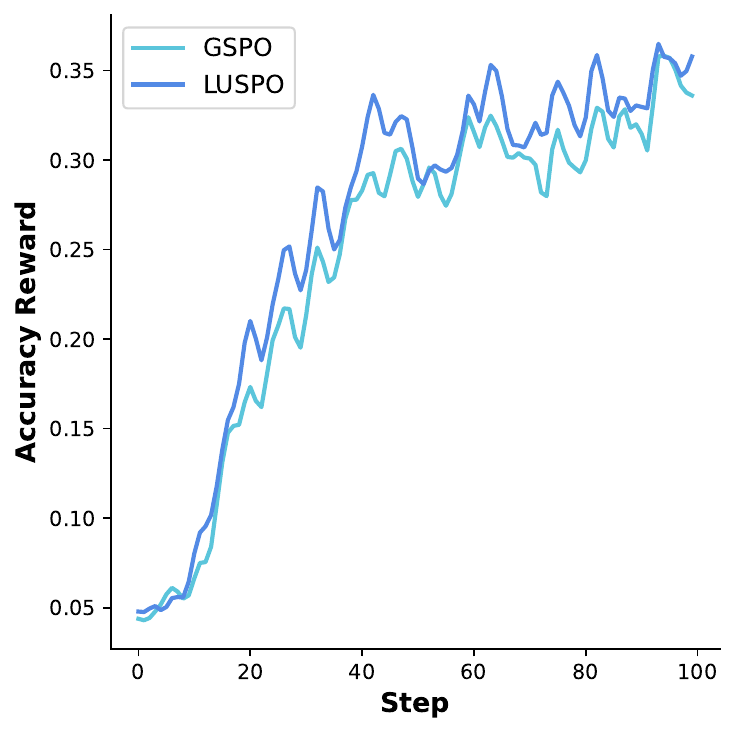}
        \caption{Qwen2.5-7B-Base}
        \label{fig:reward_qwen2.5-7b-base}
    \end{subfigure}
    \hspace{0.02\textwidth}
    \begin{subfigure}[b]{0.3\textwidth}
        \centering
        \includegraphics[width=\textwidth]{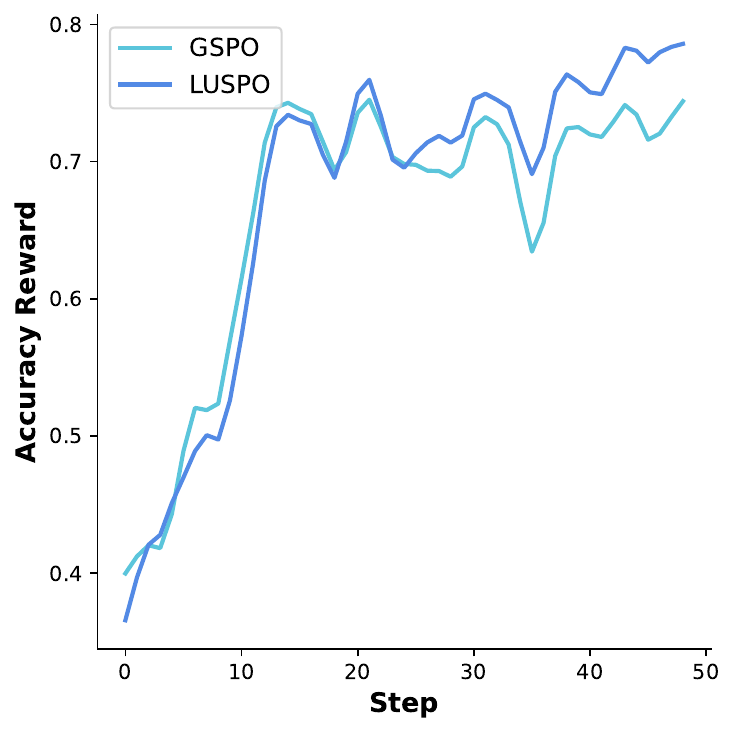}
        \caption{Qwen3-30B-A3B-Instruct}
        \label{fig:reward_qwen3-30b-a3b-instruct}
    \end{subfigure}
    \begin{subfigure}[b]{0.3\textwidth}
        \centering
        \includegraphics[width=\textwidth]{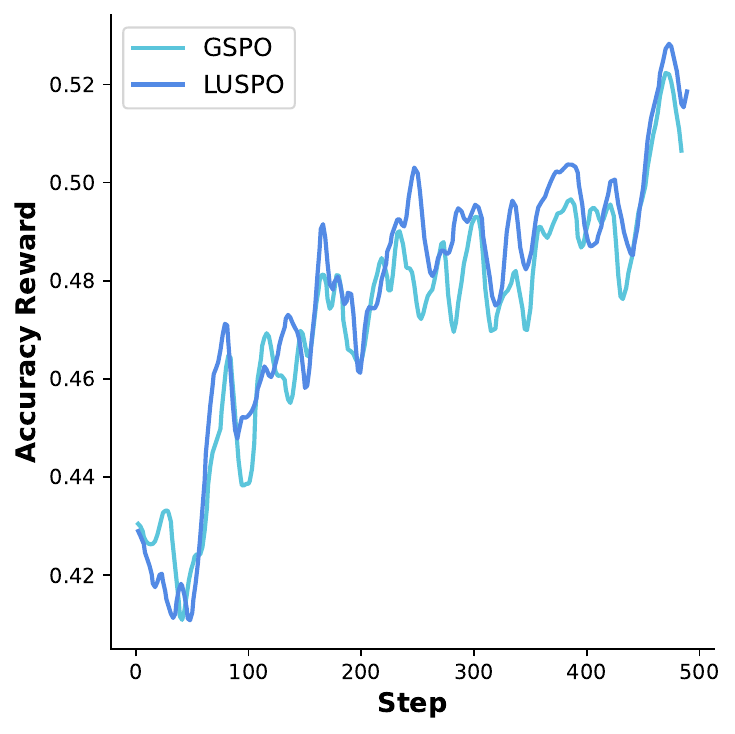}
        \caption{Qwen2.5-VL-7B-Instruct}
        \label{fig:reward_qwen2.5-vl-7b-instruct}
    \end{subfigure}
    \caption{Training curves of GSPO and LUSPO across accuracy reward.}
    \label{fig:reward}
\end{figure}


    


    

As shown in the figure \ref{fig:reward}, we present the accuracy reward of the models during training. It is evident that, under the same number of training steps, LUSPO consistently outperforms GSPO for both dense and MoE models.

This is precisely because LUSPO eliminates the inherent length bias present in GSPO. As a result, at the same number of training steps, building on the previous analysis of response length during training, LUSPO produces longer responses, giving the model a larger exploration space and making it easier to solve more complex problems.

\subsubsection*{Validation Scores}

The validation curve during training is crucial for assessing a model's generalization ability. By comparing the reward and validation curves, we can identify issues such as overfitting or underfitting. If the validation performance plateaus or begins to decline while the reward continues to improve, it indicates that the model may be memorizing the training data rather than learning general patterns.

\begin{figure}[H]
    \centering

    \begin{subfigure}[b]{0.3\textwidth}
        \centering
        \includegraphics[width=\linewidth]{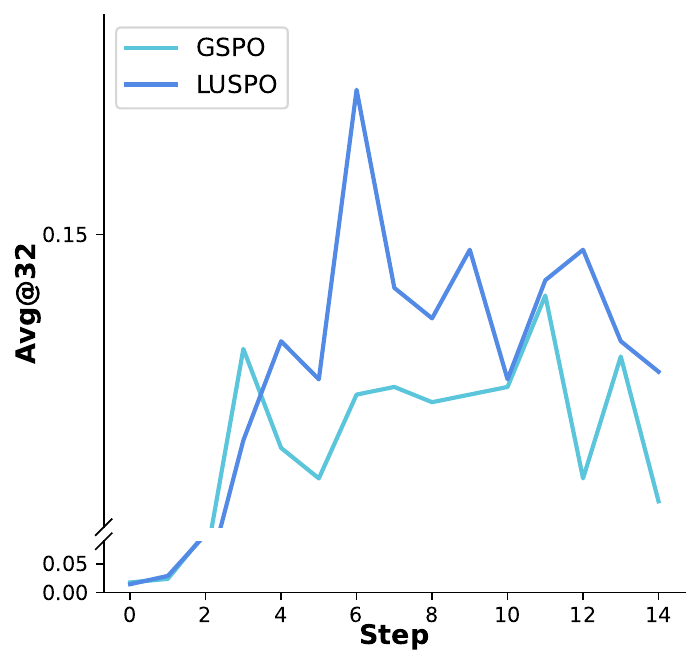}
        \caption{Qwen2.5-7B-Base}
        \label{fig:val_qwen2.5-7b-base}
    \end{subfigure}
    \hspace{0.02\textwidth}
    \begin{subfigure}[b]{0.3\textwidth}
        \centering
        \includegraphics[width=\textwidth]{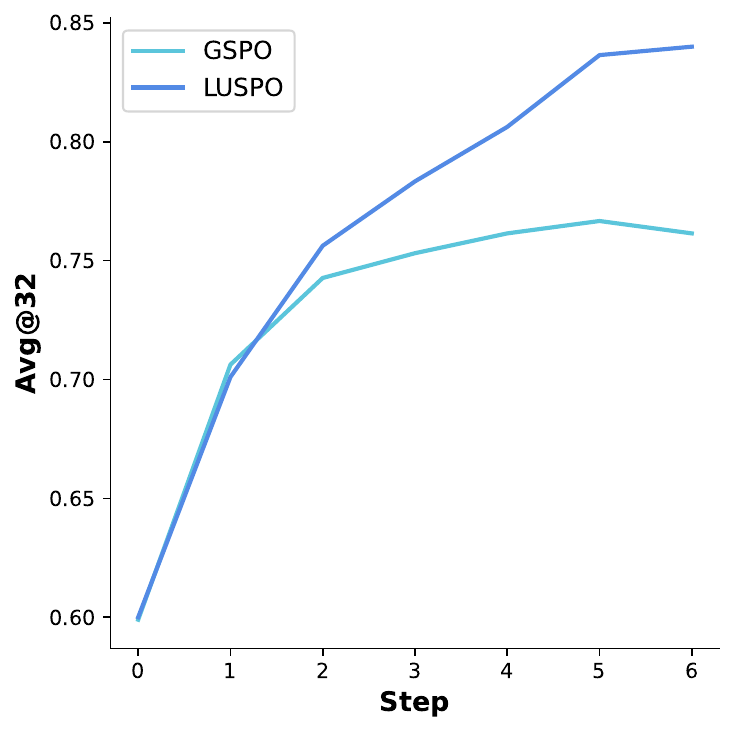}
        \caption{Qwen3-30B-A3B-Instruct}
        \label{fig:val_qwen3-30b-a3b-instruct}
    \end{subfigure}
    
    \caption{Comparison of avg@32 scores on AIME24 for LUSPO and GSPO across training.}
    \label{fig:val}
\end{figure}

During training, for Qwen2.5-7B-Base \citep{qwen2025qwen25technicalreport} and Qwen3-30B-A3B-Instruct\citep{yang2025qwen3technicalreport}, we evaluate the avg@32 metric on AIME24 \citep{aime24} every 10 steps, and plot the results as shown in the figure \ref{fig:val}.

As shown in the figure \ref{fig:val}, LUSPO not only achieves higher rewards than GSPO during training, but also demonstrates significantly improved performance on the validation set.

\subsection{Ablation Study}

\begin{table*}[!htbp]
\centering
\caption{Comparison of GSPO and LUSPO trained on ViRL39k and DAPO-MATH-17k on multimodal benchmarks.}
\label{tab:vl_benchmark_ablation}
\renewcommand{\arraystretch}{1.1} 
\setlength{\tabcolsep}{8pt}       
\resizebox{\textwidth}{!}{
\begin{tabular}{l c c c c c c c}
\toprule
\textbf{Base model + Algorithm} & MathVista-mini & MathVision & MathVerse(Vision Only) & DynaMath & WeMath & LogicVista & \textbf{Avg.} \\
\midrule

Qwen2.5-VL-7B-Instruct &  &  &  &  &  &  & \\
\hspace{1em}GSPO      & 75.3 & 28.3 & 45.2 & 24.4 & 41.3 & 47.2 & 43.6 \\
\rowcolor{rowhighlight}
\hspace{1em}LUSPO        & \textbf{75.8} & \textbf{28.8} & \textbf{49.9} & \textbf{26.3} & \textbf{45.2} & \textbf{49.4} & \textbf{45.9} \\
\rowcolor{rowaccent}
\hspace{1em}$\Delta$ \textsubscript{GSPO} & +0.5 & +0.5 & +4.7 & +2.1 & +3.9 & +2.2 & +2.3 \\
\bottomrule
\end{tabular}
}
\end{table*}

To additionally validate the robustness of our method, we also conducted training using GSPO and LUSPO on the ViRL39k and DAPO-MATH-17k, which does not cause response length collapse. Specifically, for each question in DAPO-MATH-17k, we added an extra blank image to facilitate training with VL models. The results are shown in the table \ref{tab:vl_benchmark_ablation}, and LUSPO still consistently outperforms GSPO across all five benchmarks.

\begin{figure}[H]
    \centering
    \includegraphics[width=0.4\textwidth]{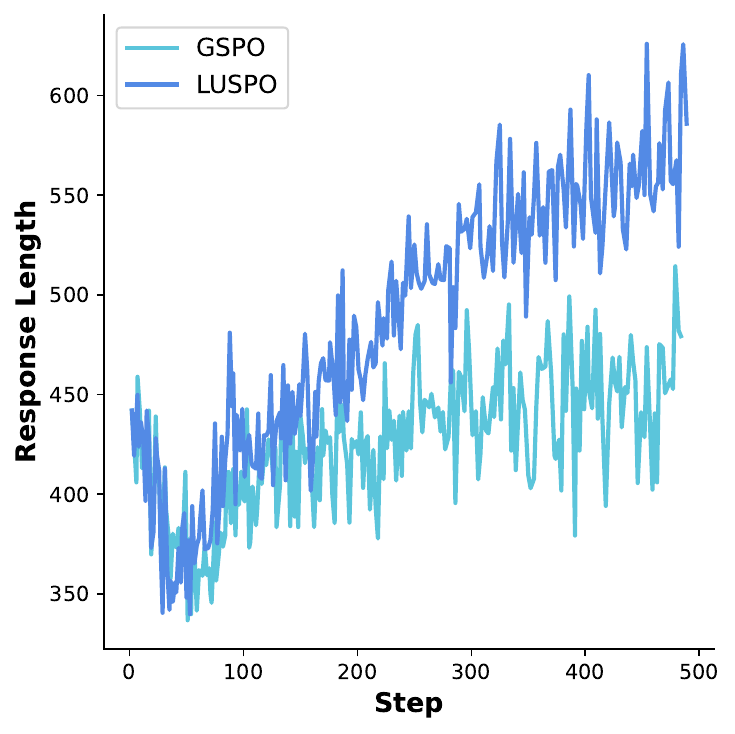}
    \caption{Response length curve during training on ViRL39k and DAPO-MATH-17k.}
    \label{fig:response_length_ablation}

    
\end{figure}

Similarly, as shown in the figure \ref{fig:response_length_ablation}, the response length during training with LUSPO is consistently higher than that with GSPO.

\section{Conclusion}
\label{section:conclusion}

In this work, based on the objective functions of GRPO and GSPO, we conduct an in-depth analysis of the reasons behind their response length bias. To address this issue and improve the stability of RLVR training for large language models, we introduce Length-Unbiased Sequence Policy Optimization (LUSPO), a novel reinforcement learning algorithm for training large language models. By applying a length-aware adjustment to sequence-level optimization, LUSPO addresses the response length bias inherent in GSPO, resulting in improved training stability and performance in both text-only and multimodal tasks. Extensive experiments on different model types and comprehensive evaluations on diverse benchmarks demonstrate the robustness and effectiveness of our method.


\clearpage
\bibliography{neurips_2025}

\begin{thebibliography}{23}
\providecommand{\natexlab}[1]{#1}
\providecommand{\url}[1]{\texttt{#1}}
\expandafter\ifx\csname urlstyle\endcsname\relax
  \providecommand{\doi}[1]{doi: #1}\else
  \providecommand{\doi}{doi: \begingroup \urlstyle{rm}\Url}\fi

\bibitem[DeepSeek(2025)]{Guo_2025}
DeepSeek.
\newblock Deepseek-r1 incentivizes reasoning in llms through reinforcement learning.
\newblock \emph{Nature}, 645\penalty0 (8081):\penalty0 633–638, September 2025.
\newblock ISSN 1476-4687.
\newblock \doi{10.1038/s41586-025-09422-z}.
\newblock URL \url{http://dx.doi.org/10.1038/s41586-025-09422-z}.

\bibitem[Shao et~al.(2024)Shao, Wang, Zhu, Xu, Song, Bi, Zhang, Zhang, Li, Wu, and Guo]{shao2024deepseekmathpushinglimitsmathematical}
Zhihong Shao, Peiyi Wang, Qihao Zhu, Runxin Xu, Junxiao Song, Xiao Bi, Haowei Zhang, Mingchuan Zhang, Y.~K. Li, Y.~Wu, and Daya Guo.
\newblock Deepseekmath: Pushing the limits of mathematical reasoning in open language models, 2024.
\newblock URL \url{https://arxiv.org/abs/2402.03300}.

\bibitem[Zheng et~al.(2025)Zheng, Liu, Li, Chen, Yu, Gao, Dang, Liu, Men, Yang, Zhou, and Lin]{zheng2025groupsequencepolicyoptimization}
Chujie Zheng, Shixuan Liu, Mingze Li, Xiong-Hui Chen, Bowen Yu, Chang Gao, Kai Dang, Yuqiong Liu, Rui Men, An~Yang, Jingren Zhou, and Junyang Lin.
\newblock Group sequence policy optimization, 2025.
\newblock URL \url{https://arxiv.org/abs/2507.18071}.

\bibitem[Zheng et~al.(2023)Zheng, Ke, Zhang, and Huang]{zheng2023clickcontrollabletextgeneration}
Chujie Zheng, Pei Ke, Zheng Zhang, and Minlie Huang.
\newblock Click: Controllable text generation with sequence likelihood contrastive learning, 2023.
\newblock URL \url{https://arxiv.org/abs/2306.03350}.

\bibitem[Yu et~al.(2025)Yu, Zhang, Zhu, Yuan, Zuo, Yue, Dai, Fan, Liu, Liu, Liu, Lin, Lin, Ma, Sheng, Tong, Zhang, Zhang, Zhang, Zhu, Zhu, Chen, Chen, Wang, Yu, Song, Wei, Zhou, Liu, Ma, Zhang, Yan, Qiao, Wu, and Wang]{yu2025dapoopensourcellmreinforcement}
Qiying Yu, Zheng Zhang, Ruofei Zhu, Yufeng Yuan, Xiaochen Zuo, Yu~Yue, Weinan Dai, Tiantian Fan, Gaohong Liu, Lingjun Liu, Xin Liu, Haibin Lin, Zhiqi Lin, Bole Ma, Guangming Sheng, Yuxuan Tong, Chi Zhang, Mofan Zhang, Wang Zhang, Hang Zhu, Jinhua Zhu, Jiaze Chen, Jiangjie Chen, Chengyi Wang, Hongli Yu, Yuxuan Song, Xiangpeng Wei, Hao Zhou, Jingjing Liu, Wei-Ying Ma, Ya-Qin Zhang, Lin Yan, Mu~Qiao, Yonghui Wu, and Mingxuan Wang.
\newblock Dapo: An open-source llm reinforcement learning system at scale, 2025.
\newblock URL \url{https://arxiv.org/abs/2503.14476}.

\bibitem[Bai et~al.(2025)Bai, Chen, Liu, Wang, Ge, Song, Dang, Wang, Wang, Tang, Zhong, Zhu, Yang, Li, Wan, Wang, Ding, Fu, Xu, Ye, Zhang, Xie, Cheng, Zhang, Yang, Xu, and Lin]{bai2025qwen25vltechnicalreport}
Shuai Bai, Keqin Chen, Xuejing Liu, Jialin Wang, Wenbin Ge, Sibo Song, Kai Dang, Peng Wang, Shijie Wang, Jun Tang, Humen Zhong, Yuanzhi Zhu, Mingkun Yang, Zhaohai Li, Jianqiang Wan, Pengfei Wang, Wei Ding, Zheren Fu, Yiheng Xu, Jiabo Ye, Xi~Zhang, Tianbao Xie, Zesen Cheng, Hang Zhang, Zhibo Yang, Haiyang Xu, and Junyang Lin.
\newblock Qwen2.5-vl technical report, 2025.
\newblock URL \url{https://arxiv.org/abs/2502.13923}.

\bibitem[Zhang and Math-AI(2024)]{aime24}
Yifan Zhang and Team Math-AI.
\newblock American invitational mathematics examination (aime) 2024, 2024.

\bibitem[Zhang and Math-AI(2025)]{aime25}
Yifan Zhang and Team Math-AI.
\newblock American invitational mathematics examination (aime) 2025, 2025.

\bibitem[Lu et~al.(2024)Lu, Bansal, Xia, Liu, Li, Hajishirzi, Cheng, Chang, Galley, and Gao]{lu2024mathvista}
Pan Lu, Hritik Bansal, Tony Xia, Jiacheng Liu, Chunyuan Li, Hannaneh Hajishirzi, Hao Cheng, Kai-Wei Chang, Michel Galley, and Jianfeng Gao.
\newblock Mathvista: Evaluating mathematical reasoning of foundation models in visual contexts.
\newblock In \emph{International Conference on Learning Representations (ICLR)}, 2024.

\bibitem[Wang et~al.(2024)Wang, Pan, Shi, Lu, Ren, Zhou, Zhan, and Li]{wang2024measuring}
Ke~Wang, Junting Pan, Weikang Shi, Zimu Lu, Houxing Ren, Aojun Zhou, Mingjie Zhan, and Hongsheng Li.
\newblock Measuring multimodal mathematical reasoning with math-vision dataset.
\newblock In \emph{The Thirty-eight Conference on Neural Information Processing Systems Datasets and Benchmarks Track}, 2024.
\newblock URL \url{https://openreview.net/forum?id=QWTCcxMpPA}.

\bibitem[Qwen(2025{\natexlab{a}})]{qwen2025qwen25technicalreport}
Qwen.
\newblock Qwen2.5 technical report, 2025{\natexlab{a}}.
\newblock URL \url{https://arxiv.org/abs/2412.15115}.

\bibitem[Qwen(2025{\natexlab{b}})]{yang2025qwen3technicalreport}
Qwen.
\newblock Qwen3 technical report, 2025{\natexlab{b}}.
\newblock URL \url{https://arxiv.org/abs/2505.09388}.

\bibitem[Liu et~al.(2025)Liu, Chen, Li, Qi, Pang, Du, Lee, and Lin]{liu2025understanding}
Zichen Liu, Changyu Chen, Wenjun Li, Penghui Qi, Tianyu Pang, Chao Du, Wee~Sun Lee, and Min Lin.
\newblock Understanding r1-zero-like training: A critical perspective.
\newblock In \emph{Conference on Language Modeling (COLM)}, 2025.

\bibitem[Gao et~al.(2025)Gao, Zheng, Chen, Dang, Liu, Yu, Yang, Bai, Zhou, and Lin]{gao2025softadaptivepolicyoptimization}
Chang Gao, Chujie Zheng, Xiong-Hui Chen, Kai Dang, Shixuan Liu, Bowen Yu, An~Yang, Shuai Bai, Jingren Zhou, and Junyang Lin.
\newblock Soft adaptive policy optimization, 2025.
\newblock URL \url{https://arxiv.org/abs/2511.20347}.

\bibitem[Wang et~al.(2025)Wang, Qu, Huang, Chu, Lin, and Chen]{vl-rethinker}
Haozhe Wang, Chao Qu, Zuming Huang, Wei Chu, Fangzhen Lin, and Wenhu Chen.
\newblock Vl-rethinker: Incentivizing self-reflection of vision-language models with reinforcement learning.
\newblock \emph{arXiv preprint arXiv:2504.08837}, 2025.

\bibitem[Sheng et~al.(2024)Sheng, Zhang, Ye, Wu, Zhang, Zhang, Peng, Lin, and Wu]{sheng2024hybridflow}
Guangming Sheng, Chi Zhang, Zilingfeng Ye, Xibin Wu, Wang Zhang, Ru~Zhang, Yanghua Peng, Haibin Lin, and Chuan Wu.
\newblock Hybridflow: A flexible and efficient rlhf framework.
\newblock \emph{arXiv preprint arXiv: 2409.19256}, 2024.

\bibitem[Loshchilov and Hutter(2019)]{loshchilov2019decoupledweightdecayregularization}
Ilya Loshchilov and Frank Hutter.
\newblock Decoupled weight decay regularization, 2019.
\newblock URL \url{https://arxiv.org/abs/1711.05101}.

\bibitem[Vaswani et~al.(2017)Vaswani, Shazeer, Parmar, Uszkoreit, Jones, Gomez, Kaiser, and Polosukhin]{vaswani2017attention}
Ashish Vaswani, Noam Shazeer, Niki Parmar, Jakob Uszkoreit, Llion Jones, Aidan~N. Gomez, {\L}ukasz Kaiser, and Illia Polosukhin.
\newblock Attention is all you need.
\newblock In \emph{Advances in Neural Information Processing Systems}, 2017.

\bibitem[Lightman et~al.(2023)Lightman, Kosaraju, Burda, Edwards, Baker, Lee, Leike, Schulman, Sutskever, and Cobbe]{lightman2023letsverifystepstep}
Hunter Lightman, Vineet Kosaraju, Yura Burda, Harri Edwards, Bowen Baker, Teddy Lee, Jan Leike, John Schulman, Ilya Sutskever, and Karl Cobbe.
\newblock Let's verify step by step, 2023.
\newblock URL \url{https://arxiv.org/abs/2305.20050}.

\bibitem[Zhang et~al.(2024)Zhang, Jiang, Zhang, Lin, Guo, Qiu, Zhou, Lu, Chang, Gao, et~al.]{zhang2024mathverse}
Renrui Zhang, Dongzhi Jiang, Yichi Zhang, Haokun Lin, Ziyu Guo, Pengshuo Qiu, Aojun Zhou, Pan Lu, Kai-Wei Chang, Peng Gao, et~al.
\newblock Mathverse: Does your multi-modal llm truly see the diagrams in visual math problems?
\newblock \emph{arXiv preprint arXiv:2403.14624}, 2024.

\bibitem[Zou et~al.(2024)Zou, Guo, Yang, Zhang, Hu, and Zhang]{zou2024dynamathdynamicvisualbenchmark}
Chengke Zou, Xingang Guo, Rui Yang, Junyu Zhang, Bin Hu, and Huan Zhang.
\newblock Dynamath: A dynamic visual benchmark for evaluating mathematical reasoning robustness of vision language models, 2024.
\newblock URL \url{https://arxiv.org/abs/2411.00836}.

\bibitem[Qiao et~al.(2025)Qiao, Tan, Dong, MinhuiWu, Sun, Song, Wang, GongQue, Lei, Zhang, et~al.]{qiao2025we}
Runqi Qiao, Qiuna Tan, Guanting Dong, MinhuiWu MinhuiWu, Chong Sun, Xiaoshuai Song, Jiapeng Wang, Zhuoma GongQue, Shanglin Lei, Yifan Zhang, et~al.
\newblock We-math: Does your large multimodal model achieve human-like mathematical reasoning?
\newblock In \emph{Proceedings of the 63rd Annual Meeting of the Association for Computational Linguistics (Volume 1: Long Papers)}, pages 20023--20070, 2025.

\bibitem[Xiao et~al.(2024)Xiao, Sun, Liu, and Wang]{xiao2024logicvistamultimodalllmlogical}
Yijia Xiao, Edward Sun, Tianyu Liu, and Wei Wang.
\newblock Logicvista: Multimodal llm logical reasoning benchmark in visual contexts, 2024.
\newblock URL \url{https://arxiv.org/abs/2407.04973}.

\end{thebibliography}
\bibliographystyle{unsrtnat}

\appendix

\end{document}